\documentstyle[aclap]{article}

\title{\vspace{-0.5in}The Effect of Pitch Accenting on Pronoun Referent
Resolution}
\author{Janet Cahn\\
Massachusetts Institute of Technology\\
Cambridge, MA 02139\\
USA\\
cahn@media.mit.edu\\}

\begin{document}

\maketitle
\vspace{-0.5in}
\begin{abstract}

By strictest interpretation, theories of both
centering and intonational meaning
fail to predict the existence of pitch accented pronominals.
Yet they occur felicitously in spoken discourse.
To explain this, I emphasize the dual functions served by pitch accents,
as markers of both propositional (semantic/pragmatic) and
attentional salience.
This distinction underlies my proposals about the
attentional consequences of pitch accents when applied to
pronominals, in particular, that  while most pitch accents may
weaken or reinforce a cospecifier's status as the center of attention,
a contrastively stressed pronominal may force a shift,
even when contraindicated by textual
features.

\end{abstract}
\section*{Introduction}

To predict and track the center of attention in discourse, theories of
centering (Grosz {\it et al.}, 1983; Brennan {\it et al.}, 1987; Grosz {\it
et al.}, 1989) and immediate focus (Sidner, 1986) rely on syntactic and
grammatical features of the text such as pronominalization and surface
sentence position. This may be sufficient for written discourse.
For oral discourse, however, we must also consider the
way intonation affects
the interpretation of a sentence, especially the cases
in which it alters the predictions of centering theories.
I investigate this via a phenomenon that, by the strictest
interpretation of either centering or intonation theories, should not occur
--- the case of {\it pitch accented pronominals}.

Centering theories would be hard pressed to predict pitch accents on
pronominals, on grounds of redundancy.  To bestow an intonational marker
of salience (the pitch accent) on a textual marker of salience (the
pronominal) is unnecessarily redundant and especially when textual
features correctly predict the focus of attention.

Intonational theories would be similarly hard pressed, but on grounds of
information quality and efficient use of limited resources. Given the
serial and ephemeral nature of speech and the limits of working
memory, it is most expedient to mark as salient the information-rich
nonpronominals, rather than their semantically impoverished pronominal
stand-ins. To do otherwise is an injudicious use of an attentional cue.

However, when uttered with  contrastive stress on the pronouns,
\newline
\hspace*{.1in}{\small \tt (1) John
 introduced Bill as \nolinebreak a \nolinebreak psycholinguist
\hspace*{.3in} and then HE insulted HIM.}
\newline
(after Lakoff, 1971) is felicitously understood to mean that after a
slanderous introduction, Bill retaliated in kind against John.

What makes {\tt (1)} felicitous is that the pitch accents on the pronominals
contribute attentional information that cannot be gleaned from text alone.
This suggests an attentional component to pitch accents, in addition to
the propositional component explicated in Pierrehumbert and Hirschberg
(1990).
In this paper, I combine their account of
pitch accent semantics with Grosz,  Joshi and Weinstein's (1989) account
of centering to yield insights into the phenomenon
of pitch accented pronominals, and the attentional consequences of pitch
accents in general.
The relevant claims in PH90 and GJW89 are reviewed in the next two sections.

\section* {Pitch accent semantics}

A {\it pitch accent} is a distinctive intonational contour
applied to a word to convey sentential stress (Bolinger, 1958;
Pierrehumbert, 1980). PH90 catalogues six pitch accents, all combinations
of high (H) and low (L) pitch targets, and structured as a main tone
and an optional leading or trailing tone. The form of
the accent --- L, H, L+H or H+L --- informs about the operation
that would relate the salient item to the
{\it mutual beliefs}\footnote{{\it Mutual beliefs:}
propositions expressed or implied by the discourse, and which all
conversants believe each other to accept as true and relevant
same (Clark and Marshall, 1981).} of the conversants; the main tone either
commits (H*) or
fails to commit (L*) to the salience of the proposition itself, or the
relevance of the operation.

\begin{itemize}

\item H* {\it predicates} a proposition as mutually believed, and proclaims
its {\it addition} to the set of mutual beliefs; L* {\it fails to
predicate} a proposition as mutually believed. As PH90 points out,
failure to predicate has contradictory sources: the proposition
has already been predicated as mutually believed; or, the speaker, but
not the hearer, is prevented from predication (perhaps by social
constraints); or the speaker actively believes the salient proposition to be
false.

\item H+L evokes an {\it inference path}. H*+L commits to the existence
of inference path that would support the proposition as mutually
believed, indicates that it can
be found or derived from the set of mutual beliefs; H+L* conveys
uncertainty about the existence of such a path.

\item L+H evokes a {\it scale or ordered set} to which the accented constituent
belongs: L+H* commits to the salience of the scale, and is typically used
to convey contrastive stress; L*+H also evokes a scale but fails to commit to
its salience, e.g., conveying uncertainty about the salience of the
scale with regard to the accented constituent.

\end{itemize}

\section* {Centering structures and operations}

To explain how speakers move an entity in and out of the center of [mutual]
attention, GJW89 formalizes attentional operations with two computational
structures --- the {\it forward-looking center list} (Cf) and the {\it
backward-looking center} (the Cb). Cf is a partially ordered list of
centering candidates;\footnote{For simplicity's sake, we assume the items
in Cf to be words and phrases; in actuality, they may be
nonlexical representations of concepts, or some hybrid of lexical,
conceptual and sensory data.} the Cb, at the head of Cf, is the current
center of attention.

After each utterance, one of three operations are possible:

\begin{itemize}

\item The Cb retains both its position at the head of Cf and its status as
the Cb; therefore it {\it continues} as the center in the next utterance.

\item The Cb {\it retains} its centered status for the current utterance
but its rank is lowered --- it no longer resides at the head of Cf and
therefore ceases to be the center in the next utterance.

\item The Cb loses both its centered status and ranking in the current
utterance as attention {\it shifts} to a new center.

\end{itemize}

In addition, GJW89 constrains pronominalization such that no element in an
utterance can be realized as a pronoun unless the Cb is also realized as a
pronoun, and imposes a preference ordering for operations on Cf, such that
the least reordering is always preferred. That is, a sequence of
continuations is preferred over a sequence of retentions, which is
preferred over a sequence of shifts.

\section* {When intonation and centering collide}

My synthesis of the claims in PH90 and GJW89  produces an
attentional interpretation of pitch accents,
modeled by operations on Cf, and  derived for
each accent from their corresponding propositional effect as
described in PH90.

The corollaries for pitch accented pronominals are:
(1) when a pitch accent is applied to a pronominal, its main effect
is attentional, on the order of items in Cf;  (2)  the obligation to accent a
pronominal
for attentional reasons depends on the variance between what the
text  predicts and what the speaker would like to assert
about the order of items in Cf.

These hypotheses arise from the following chain of assumptions:

\vspace*{2pt}
\noindent{\em (1) To analyze the effects of pitch accents on pronominals, it is
necessary to distinguish between attentional and propositional salience.}
Attentional salience  measures
the degree to which an item is salient, expressible as
a partial ordering,  e.g., its ranking in Cf.   It is a quantitative
feature.
In contrast, propositional salience, addressing an item's status in relation
to mutual beliefs, is qualitative.
It is  calculated through
inference chains that link semantic and pragmatic propositions.

Both attentional (Cf) and propositional (mutual beliefs) structures
are updated throughout. However, unlike attentional structures
which are ephemeral in various time scales and empty at the end of the
discourse (Grosz and Sidner, 1986), mutual beliefs persist throughout the
conversation, preserving at the end the semantic and pragmatic outcome of
the discourse.

In addition, while propositions can be excluded from the mutual beliefs
because they fail to meet some inclusion criterion, no lexical denotation
is excluded from Cf regardless of its propositional value. This is because
the  salience most relevant to the attentional state is the proximity of
a discourse entity to the head of Cf --- the closer it is, the more it
is centered and therefore, attentionally salient.

\vspace*{2pt}
\noindent
{\em (2) Pitch accents on pronominals are primarily interpreted for what
they say about attentional salience.} One determiner of whether attentional
or propositional effects are dominant is the type of information provided
by the accented constituent. Because nonpronominals contribute
discourse content, pitch accented nonpronominals are mainly interpreted
with respect to the mutual beliefs, that is, for their propositional
content. However, pronominals, with little intrinsic semantics, perform
primarily an attentional function. Therefore pitch accented pronominals are
mainly interpreted with respect to Cf, for their attentional content.

\vspace*{2pt}
\noindent
{\em (3) The specific attentional consequences of each pitch
accent on pronominals can be extrapolated by analogy from
the propositional interpretations in PH90, by replacing
mutual beliefs with Cf as the salient set.} Thus,

\begin{itemize}

\item H* indicates instantiation of the pronominal's cospecifier as the
Cb, while L* fails to instantiate it as the Cb;

\item The partially ordered set (salient scale) invoked by L+H
is Cf;

\item The inference path evoked by H+L is, for attentional purposes, a
traversal of Cf.

\end{itemize}

\noindent
{\em (4) And therefore, the attentional effect of pitch accents can be
formally expressed as an effect on the order of items in Cf.}

\vspace*{2pt}
{}From these assumptions, I derive the following attentional consequences
for pitch accented pronominals:

\vspace*{-7pt}
\begin{itemize}

\item Only one pitch accent, L+H*, selects a Cb other than that predicted by
centering theory and thereby reorders Cf.

\item L*+H appears to support an impending reordering but does not compel
it.

\item By analogy, the remaining pitch accents, seem to
either weaken or strengthen the current center's Cb status, but
do not force a reordering.

\end{itemize}

\section*{Availability of cospecifiers}

The attentional interpretations are constrained by what has
been mutually established in the prior discourse, or is situationally
evident.
Therefore, while contrastive stress may be
mandated when grammatical features select the wrong cospecifier, the
accenting is only felicitous {\it when there is an alternate referent
available.}

For example, in
\newline
\hspace*{.2in}{\small \tt (2) John introduced Bill as a psycholinguist
\hspace*{.3in} and then he$_{L+H*}$ insulted him.}
\newline
L+H* indicates that {\tt he} no longer cospecifies with {\tt John}.
If the hearer is hasty, she might
select {\tt Bill} as the new Cb. However, this is not borne out by the
unaccented {\tt him}, which continues to cospecify with  {\tt Bill}. Since {\tt
he} and
{\tt him} cannot select the same referent,
{\tt he} requires a cospecifier that is neither {\tt John}
nor {\tt Bill}.  Because, the utterance itself does not provide a
any other alternatives,
{\tt he$_{L+H*}$}
is only felicitous (and coherent) if an alternate cospecifier has been placed
in Cf by prior discourse,  or by
the speaker's concurrent
deictic gesture towards a discourteous male.

\section*{Conclusion and Future Work}

By combining Pierrehumbert and Hirschberg's (1990) analysis
of intonational meaning with Grosz, Joshi and Weinstein's (1989)
theory of centering in discourse,  the attentional affect of pitch accents
becomes evident, and the paradox of pitch accented pronominals
unravels.  My goal here is  to develop an analysis and a line of inquiry
and to suggest
that my derivative
claims are plausible, and even
extensible to an attentional analysis of pitch accents on
nonpronominals.
The proof, of course, will come from investigation
by multiple means --- constructed examples (e.g., Cahn, 1990), computer
simulation, empirical analysis of speech data (e.g., Nakatani, 1993),
and psycholinguistic experiments.

\section*{References}

Dwight Bolinger. {A Theory of Pitch Accent in English}. {\em Word},
14(2--3):109--149, 1958.

Susan~E. Brennan, Marilyn~W. Friedman, and Carl~J. Pollard. {A Centering
Approach to Pronouns}. {\em Proceedings of the 25th Conference of the
Association for Computational Linguistics}, 1987.

Janet Cahn. {The Effect of Intonation on Pronoun Referent Resolution.}
{\em Draft}, 1990.  Available as:
Learning and Common Sense TR 94-06, M.I.T. Media Laboratory.

Herbert~H. Clark and Catherine~R. Marshall. {Definite Reference and Mutual
Knowledge}.  In Webber, Joshi and Sag, editors, {\em Elements of Discourse
Understanding}. Cambridge University Press, 1981.

Barbara Grosz, Aravind~K. Joshi, and Scott Weinstein. Providing a unified
account of definite noun phrases in discourse. {\em Proceedings of the 21st
Conference of the Association for Computational Linguistics}, 1983.

Barbara~J. Grosz, Aravind~K. Joshi, and Scott Weinstein. Towards a
Computational Theory of Discourse Interpretation. {\em Draft},
1989.

Barbara~J. Grosz and Candace~L. Sidner. {Attention, Intentions, and the
Structure of Discourse}. {\em Computational Linguistics}, 12(3):175--204,
1986.

George Lakoff. Presupposition and relative well-formedness.  In Danny D.
Steinberg and Leon A. Jakobovits, editors, {\em Semantics: An
Interdisciplinary Reader in Philosophy, Linguistics and Psychology},
Cambridge University Press, 1971.

Christine Nakatani. Accenting on Pronouns and Proper Names in Spontaneous
Narrative. {\em Proceedings of the European Speech Communication Association
Workshop on Prosody}, 1993.

Janet~B. Pierrehumbert. {\em The Phonology and Phonetics of English
Intonation}. Ph.D. thesis, Massachusetts Institute of Technology, 1980.

Janet~B. Pierrehumbert and Julia Hirschberg. {The Meaning of Intonation
Contours in the Interpretation of Discourse}. In Philip~R. Cohen, Jerry
Morgan, and Martha~E. Pollack, editors, {\em Intentions in Communication},
MIT Press, 1990.

Candace~L. Sidner. {Focusing in the Comprehension of Definite Anaphora}. In
Barbara~J. Grosz, Karen Sparck-Jones, and Bonnie~Lynn Webber, editors, {\em
Readings in Natural Language Processing}, Morgan Kaufman
Publishers, Inc., 1986.

\end{document}